\documentclass[10pt,twocolumn,letterpaper]{article}

\usepackage{wacv}
\usepackage{times}
\usepackage{epsfig}
\usepackage{graphicx}
\usepackage{amsmath}
\usepackage{amssymb}
\usepackage{multirow}

\wacvfinalcopy 

\def\wacvPaperID{****} 

\ifwacvfinal\fi
\begin{document}

\title{Stacked Adversarial Network for Zero-Shot Sketch based Image Retrieval}

\author{
Anubha Pandey$^1$ \\
{\tt\small anubhap93@gmail.com}
\and
Ashish Mishra$^1$ \\
{\tt\small mishra@cse.iitm.ac.in}
\and
Vinay Kumar Verma$^2$ \\
{\tt\small vkverma@iitk.ac.in}
\and
Anurag Mittal$^1$ \hspace{2cm} Hema A. Murthy$^1$   \\
{\tt\small amittal@cse.iitm.ac.in \hspace{0.5cm} \tt\small hema@cse.iitm.ac.in}\\
$^1$Indian Institute of Technology Madras \hspace{2cm} $^2$Indian Institute of Technology Kanpur
}

\maketitle
\ifwacvfinal\thispagestyle{empty}\fi

\begin{abstract}
Conventional approaches to Sketch-Based Image Retrieval (SBIR) assume that the data of all the classes are available during training. The assumption may not always be practical since the data of a few classes may be unavailable, or the classes may not appear at the time of training. Zero-Shot Sketch-Based Image Retrieval (ZS-SBIR) relaxes this constraint and allows the algorithm to handle previously unseen classes during the test. This paper proposes a generative approach based on the Stacked Adversarial Network (SAN) and the advantage of Siamese Network (SN) for ZS-SBIR. While SAN generates a high-quality sample, SN learns a better distance metric compared to that of the nearest neighbor search. The capability of the generative model to synthesize image features based on the sketch reduces the SBIR problem to that of an image-to-image retrieval problem. We evaluate the efficacy of our proposed approach on TU-Berlin, and Sketchy database in both standard ZSL and generalized ZSL setting. The proposed method yields a significant improvement in standard ZSL as well as in a more challenging generalized ZSL setting (GZSL) for SBIR.
\end{abstract}

\section{Introduction}

The standard approaches for retrieving related information from a  huge database of images are either based on a query image or query text. Retrieval of images using an image-based query is relatively easy compared to that of image retrieval using text-based queries. Text-based queries can be ambiguous, incomplete, and language-dependent. Recent research has shown that instead of text descriptions, sketches can be used as a query. It is more convenient to use sketches as queries since shapes are easy to remember than the textual description. Image retrieval using sketch-based queries is referred to as sketch-based image retrieval (SBIR).
\cite{conf/iccv/SBIR1,sia2,conf/cvpr/SBIR2,conf/cvpr/SBIR15}.

SBIR aims to retrieve the images that belong to a class using a set of query sketches from the same class. Freehand sketches may magnify the cross-domain discrepancy between sketches and the real-world images as they can vary significantly across persons depending upon the salient features of the image that a person wants to emphasize. In order to make retrieval robust, sketches and their corresponding images are projected to a common subspace \cite{berlin, sarthak,sketchy}. The major issue with this approach is that the method fails to generalize for the test data under the unavailability of accurate sketches, and its performance on unseen classes is poor. 

To address these issues recently, Dey et al.\cite{doodle}, Dutta et al. \cite{akata_cvpr19}, Verma et al. \cite{vinay_cvprw19} Pandey et al. \cite{anubha_iccvw19}, Shen et al.\cite{imagehashing}, and Yelamarthi et al.\cite{ashisheccv2018} proposed SBIR in Zero-Shot framework(ZS-SBIR).In ZS-SBIR, the training and testing classes are mutually exclusive. Shen et al. \cite{imagehashing} in their proposed ZSIH approach, combined zero-shot learning and sketch-based image retrieval using a cross-modal hashing scheme. Dey et al. \cite{doodle} proposed a ZS-SBIR framework that learns a common embedding space for both the sketch and image domains. \cite{doodle,akata_cvpr19,imagehashing} use sketch class descriptions\cite{glove} as side information along with sketch features for establishing the semantic relationship between the image feature space and sketch feature space. In contrast, Yelamarthi et al.\cite{ashisheccv2018} proposed two similar autoencoder-based generative models, CAAE(Conditional Adversarial Autoencoder) \cite{AAE} and CVAE(Conditional Variational Autoencoder)\cite{conf/nips/CVAE} for zero-shot SBIR without using any side information. One of the major shortcomings of the ZSIH\cite{imagehashing} and Doodle to search \cite{doodle} is that they require sketch class descriptions as side information for learning semantics between the sketches and images. Due to the explosive growth of new categories, it is not practically possible to get class descriptions for every new class. We propose a generative model for SBIR in the zero-shot framework, which shows a significant improvement without using any side information among all the state-of-the-art methods for both the datasets Sketchy\cite{sketchy} and Berlin\cite{berlin}. 

Zero-shot learning is categorized into two settings based on the test data. One is standard zero-shot learning(ZSL), which assumes that the seen and unseen classes are mutually exclusive, and the test data comes only from the unseen classes\cite{conf/nips/DeviSE, ConSE}. The other one is generalized zero-shot learning(GZSL), which assumes that the test data may belong to both the seen and unseen classes\cite{akata2015evaluation, SAE2017,verma2017simple}. GZSL setting is more challenging as compared to the standard ZSL setting. So it is observed that most of the existing approaches are biased towards the seen classes for the GZSL setting. The prior works for ZS-SBIR, ZSIH\cite{imagehashing}, CVAE\cite{ashisheccv2018}, Doodle to search \cite{doodle}, JGAN \cite{anubha_iccvw19} and GZS-SBIR \cite{vinay_cvprw19} have shown experiments only for standard ZSL setting, whereas our proposed model has shown competitive performances on both the ZSL and GZSL settings. 

\begin{figure}[t]
    \centering
    \includegraphics[scale=0.37]{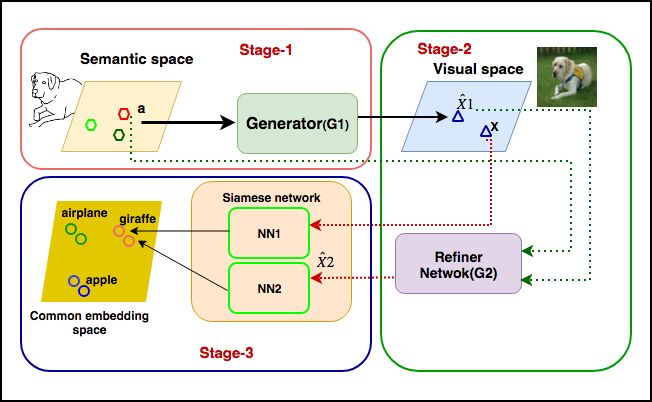}
    \caption{Overview of the proposed approach}
    \label{fig:overview}
\end{figure}

In this paper, we propose a multistage generative model for the sketch-based image retrieval task in a zero-shot setting. The model is inspired by the StackGan architecture \cite{stackgan}.  The output of the multistage model is fed to the Siamese-Network(SN) \cite{siamese} to learn a better embedding and reduce the Hubness problem \cite{hubness}. We believe that using multiple stages of GAN, we can generate refined features that are more close to the original image feature space. Further, using Siamese Network\cite{siamese}, we project the generated and real image features into another space where they are more discriminative. The Siamese network uses Contrastive loss function to distinguish between the given pair of generated and real image features in the projected space. This approach helps to reduce the ZS-SBIR problem into multiple subproblems: Stage1- Projection of sketch features to image domain, Stage2- Refinement of generated image features and Stage3- Generation of more distinctive features using Siamese Network. The generative nature of the model enables the synthesis of the pseudo labeled image instances for unseen classes based on sketch features. This approach converts the zero-shot SBIR (ZS-SBIR) problem into a conventional image-to-image retrieval problem. The overview of the proposed method is shown in Figure \ref{fig:overview}.
Our contribution is summarized below:
  \begin{itemize}
      \item We propose a multi-stage GAN based generative model for zero-shot setting that transforms the zero-shot sketch-based image retrieval (ZS-SBIR) problem to a conventional image-to-image retrieval problem.
      \item  We propose to use a Maximum Mean Discrepancy (MMD)loss\cite{mmd} in GAN \cite{conf/nips/GAN} it helps to distinguish between the pairs of real and generated features of images of different classes.
      \item Unlike the previous approaches for ZS-SBIR \cite{imagehashing,ashisheccv2018}  that performs a nearest neighbor search in the image space, we use a Siamese Network based on the max-margin loss to learn a better metric for the similarity measured in the projected space, inspired by the prior work Qi et.al\cite{sia2}.
      \item Our method yields significantly better results in both the standard and generalized zero-shot setting without using any side information (e.g., word2vec based attributes of the classes\cite{word2vec,glove}), as compared to \cite{ashisheccv2018}. 
  \end{itemize}  
  
\section{Related Work}
In this section, we briefly describe the existing techniques for both SBIR and zero-shot learning. Free hand-drawn sketches fail to capture the complete information of the images; this causes a significant cross-domain gap between the sketch and the image feature space. SBIR tries to learn a shared representation for both the sketches and the images to mitigate the domain gap between the two different spaces. The traditional methods in SBIR, such as \cite{liu,sketchy, journals/corr/SBIR13}, used hand-crafted descriptors of sketches and images for retrieval. The conventional deep learn frameworks of SBIR try to project features of sketches and images into a common subspace such that the sketches and images of the same class project close to each other, while the projection of sketches and images of different classes are distant. These projected features are used in the retrieval task. Qi et.al\cite{sia2} used Siamese architecture and Sangkloy et.al \cite{sketchy} used triplet ranking loss for coarse-grained SBIR. Liu et.al\cite{liu} proposed a semi-heterogeneous deep architecture for extracting the binary codes from the sketches and the images, which can be trained in an end-to-end fashion for the coarse-grained SBIR task.

Existing SBIR approaches  \cite{liu,sia2,sketchy,journals/corr/SBIR13,journals/ijcv/CNNinSKETCH} do not generalize in terms of learning the mapping for unseen sketches, and the corresponding classes. Similarly, state-of-the-art methods for SBIR work well for already seen classes, whereas for any new class, they fail to retrieve the same class images. The capability of zero-shot learning (ZSL) to classify an unseen class example at the test time has received significant attention  \cite{ akata2015evaluation, conf/nips/DeviSE,SAE2017, ConSE, conf/nips/CMT}. ZSL aims to recognize instances of unseen classes by a transfer of semantic information from seen to unseen classes. There are primarily two different approaches to ZSL.

\begin{figure*}[htb!]
\begin{center}
\includegraphics[scale=0.32]{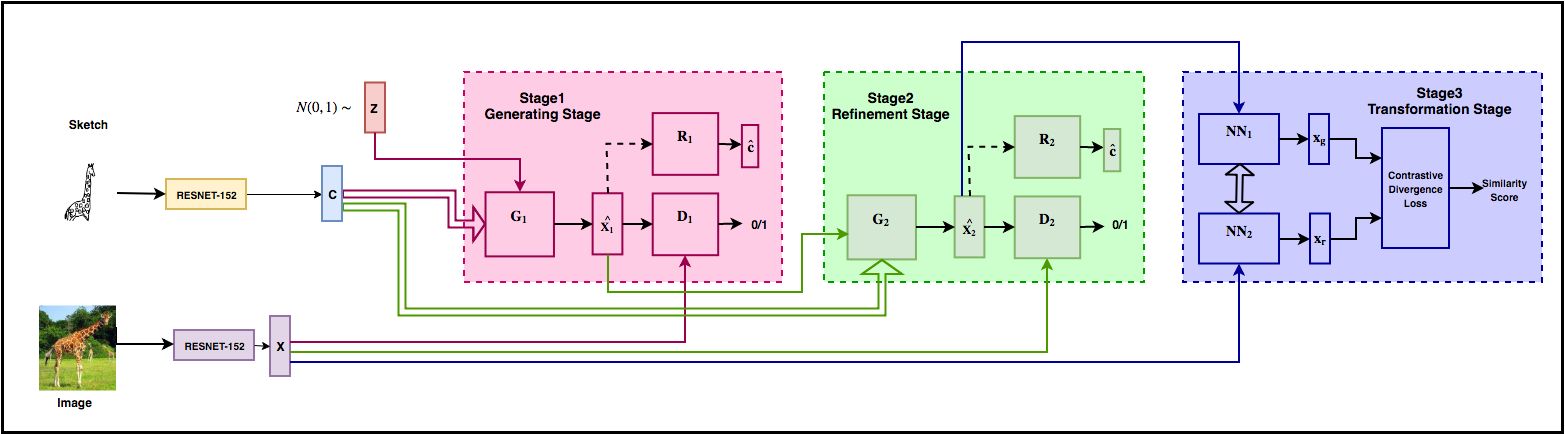}
\end{center}
\caption{The Training pipeline of our proposed model. The features for images \textbf{x} and sketches \textbf{c} are extracted using same pretrained ResNet-152 on ImageNet-1000 dataset.}
\label{fig:main}
\end{figure*}
The first type is embedding-based ZSL. Embedding based approaches \cite{akata2015evaluation,mishracvpr,mishrawacv,ConSE,verma2017simple,vinayaaai,xu2017transductive} address this issue by learning the interaction between visual space and semantic or class attributes space. Based on the direction of the embedding function, they are divided into three sub-categories. The first one learns the embedding from visual space to semantic space. The second approach learns the embedding from the semantic space to the visual space. Both of these approaches suffer from the hubness problem \cite{hubness},i.e., a small number of objects (hubs) may occur as the nearest neighbor of many categories, resulting in the diminishing of the nearest neighbor method. To address this issue, the third type of approach learns a bilinear embedding function to project both the visual features and class prototypes or semantic features into a shared latent space. This suffers from the domain-shift problem. 

The second type of approach is the synthesis-based ZSL \cite{guo2017synthesizing,vinaycvpr,mishrawacv,verma_meta,verma2017simple,xian2018feature}. These are recent generative approaches to zero-shot learning. Synthesis-Based ZSL converts the zero-shot learning problem to the traditional supervised learning problem by synthesizing pseudo-labeled data based on class-prototype or semantic description of unseen classes.  

Recently \cite{doodle,akata_cvpr19,vinay_cvprw19,anubha_iccvw19,imagehashing} and \cite{ashisheccv2018} have proposed an approach for sketch-based image retrieval in the zero-shot framework.  \cite{imagehashing} proposed a hashing based model for ZS-SBIR, \cite{doodle} has proposed to learn a joint distribution between sketch and image domain, and \cite{ashisheccv2018} proposed two models, one is based on conditional variational auto-encoder (CVAE). The second is based on conditional adversarial auto-encoder(CAAE) for the ZS-SBIR task. Also, \cite{doodle, akata_cvpr19} and \cite{imagehashing} use sketch class description as an additional information whereas \cite{ashisheccv2018} does not use any side information to train the model. In this paper, we propose a multi-stage conditional generative adversarial network inspired by stackGan architecture \cite{stackgan} followed by a Siamese network for matching. Our model does not use any additional information other than sketch features for zero-shot training similar to \cite{ashisheccv2018}.

\section{Proposed Approach}
\subsection{Zero-Shot SBIR (ZS-SBIR)}
In the zero-shot setting, we partition the dataset into two mutually exclusive sets based on sketch classes: Seen classes(S) and Unseen classes(U) i.e., $S\cap U=\phi$. The train data belongs to the Seen Classes(S). In the Standard ZSL setting, test data belongs to the Unseen Classes(U), and in the Generalized ZSL setting, test data belongs to both the Seen and Unseen Classes. 
The objective of zero-shot learning is to train a model that generalizes well for unseen class sketches as well. The mathematical formulation and notations of the ZS-SBIR are given below:

Let $A=\{(\mathbf{x_i}^{s},\mathbf{x_i}^{im},\mathbf{y_i})|\mathbf{y_i} \in \mathcal{Y}\}$ be the triplet of sketch, image, and the class label. Here $\mathcal{Y}$ is the set of all class labels. We partition the class labels in the data into $\mathcal{Y}_{train}$ and $\mathcal{Y}_{test}$ for the train and test respectively. Let $A_{tr}=\{\mathbf{x_i}^{s},\mathbf{x_i}^{im},\mathbf{y_i}|\mathbf{y_i} \in Y_{train}\}$ and $A_{te}=\{\mathbf{x_i}^{s},\mathbf{x_i}^{im},\mathbf{y_i}|\mathbf{y_i} \in Y_{test}\}$ be the partition of $A$ into train and test sets. We denote sketch feature $\mathbf{x}^{s}$ with $\mathbf{c}$ and image feature $\mathbf{x}^{im}$ with $\mathbf{x}$ through out this paper for convenience. Another assumption for the ZS-SBIR is $A_{tr}\cap A_{te}=\phi$.

The overall architecture of the proposed system consists of three stages, as described below:
\subsection{Stage-1}
The first module consists of a Conditional Generative Adversarial Network (CGAN). It takes sketch features and a random vector from the unit Gaussian distribution as input and generates the corresponding class image features. The main task of this module is to generate the image features, conditioned on the same class sketch feature. We call it a generator module.

This module is composed of a generator $G_1:C \times Z \rightarrow X$ parameterized by $\mathbf{\theta_{G_1}}$, a discriminator $D_1:X \rightarrow [0,1]$ parameterized by $\mathbf{\theta_{D_1}}$ and a regressor $R_1:X
\rightarrow C $ parameterized by $\mathbf{\theta_{R_1}}$.
Where C is a set of conditional attributes(sketch features) and Z is a set of random vectors sampled from a unit Gaussian. Generator $G_1$ takes as input a sketch feature $\mathbf{c}$ and random vector $\mathbf{z}$ which is sampled from $ \mathcal{N}(0,1)$ and generates the image feature $\mathbf{\hat{X_1}}$ of the same class as that of the sketch. Discriminator $D_1$ takes input as real image feature $\mathbf{X}$ or generated image feature $\mathbf{\hat{X_1}}$ and attempts to distinguish between real features, and synthesized features. Regressor $R_1$ acts as a regularizer for the generator $G_1$, where it tries to reconstruct the original sketch feature from the generated image feature $\mathbf{\hat{X_1}}$. Regressor $R_1$ helps the generator $G_1$ to generate more discriminative and realistic image features. The loss functions used are:
\begin{equation}\label{eq:supervised1}
    L_{rec}(\mathbf{\theta_{G_1}})=||\mathbf{X}-G_1(\mathbf{c,z;\theta_{G_1}})||_{2}
\end{equation}
\begin{equation}\label{eq:adversarial1}
    \begin{split}
        L_{adv}(\mathbf{\theta_{G_1},\theta_{D_1}})= & log(D_1(\mathbf{X;\theta_{D_1}})) \\ & - log(1-D_1(G_1(\mathbf{c,z;\theta_{G_1}}))
    \end{split}
\end{equation}
\begin{equation}\label{eq:reconstruction1}
    L_{reg}(\mathbf{\theta_{G_1},\theta_{R_1}})=||\mathbf{c}-R_1(G_1(\mathbf{c,z;\theta_{G_1});\theta_{R_{1}}})||
\end{equation}
Here $L_{rec}$ is the reconstruction loss, $L_{adv}$ is the adversarial loss and $L_{reg}$ is the regularizer loss. The overall GAN loss is given as :
\begin{equation}\label{eq:joint1}
    L_{GAN_1}(\mathbf{\theta_{G_1},\theta_{D_1},\theta_{R_1}})=L_{rec}+\alpha *L_{adv}+\beta * L_{reg}
\end{equation}
Here $\alpha$ and $\beta$ are hyper-parameters. In the proposed approach, instead of pure adversarial loss (Equation \ref{eq:adversarial1}), we include the supervised mean square error loss (Equation \ref{eq:supervised1}). Empirically we have found that the joint loss given in Equation \ref{eq:joint1} shows better results than the adversarial loss. 

\subsection{Stage-2}
This module uses an architecture similar to the Stage-1, but the task is to refine the features generated in Stage-1. The StackGan architecture \cite{stackgan} inspires the combination of Stage-1 and Stage-2, where the first GAN learns to generate high-level features, and the second GAN learns to generate low-level features. Because of the multiple stage refinement, StackGan generates more realistic images as compared to a single GAN. 
The Generator $G_2$ takes the generated feature $\mathbf{\hat{X_1}}$ from the Stage-1 and its corresponding attribute $\mathbf{c}$ as input, and generates the refined feature  $\mathbf{\hat{X_2}}$. The Discriminator $D_2$ takes the real image features $\mathbf{X}$ and the generated image features $\mathbf{\hat{X_2}}$ as input and classifies them as synthetic or real. The Regressor $R_2$ acts as a regularizer by reconstructing the original attribute using the generated features $\mathbf{\hat{X_2}}$. This regularization step helps $G_2$ to generate more discriminative features that are close to that of the actual image. The loss functions used are:
\begin{equation}\label{eq:supervised2}
    L_{rec}(\mathbf{\theta_{G_2}})=||\mathbf{X}-G_2(\mathbf{c,\hat{X_1};\theta_{G_2}})||
\end{equation}
\begin{equation}\label{eq:adversarial2}
    \begin{split}
    L_{adv}(\mathbf{\theta_{G_2},\theta_{D_2}})= & log(D_2(\mathbf{X;\theta_{D_2}})) \\ & \quad - log(1-D_2(G_2(\mathbf{c,\hat{X_1};\theta_{G_2}}))
    \end{split}
\end{equation}
\begin{equation}\label{eq:reconstruction2}
    L_{reg}(\mathbf{\theta_{G_2},\theta_{R_2}})=||\mathbf{c}-R_2(G_2(\mathbf{c,\hat{X_1};\theta_{G_2});\theta_{R_2}})||
\end{equation}

We further add a Maximum Mean Discrepancy loss(MMD)\cite{mmd} in the generator $G_2$. The MMD loss is a kernel-based distance function between pairs of synthesized and real samples. Using MMD loss, we project both the synthesized and real image features in a high dimensional space using a kernel function and try to preserve the property of the image class. MMD loss also acts as a regularizer for generator $G_2$ to generate more discriminative and similar features to the original class image features. We compute MMD loss between generated image features $\hat{\mathbf{X_2}}$, and real image features $\mathbf{X}$. 
Assume $\mathbf{x}$ is real image feature and $\mathbf{\hat{x}}$ is the generated image feature The overall MMD loss for all N training samples is defined as : 
\begin{equation}\label{eq:mmd_img}
 \begin{split}
 L_{Img}^{mmd}\left(\mathbf{x,\hat{x}}\right)=& \sum_{j=1}^{j=N}\sum_{j'=1}^{j'=N} k(\mathbf{x_j,x_{j'}}) - 2 \sum_{j=1}^{j=N}\sum_{i=1}^{i=N} k(\mathbf{x_j,\hat{x}_i})\\ & +\sum_{i=1}^{i=N}\sum_{i'=1}^{i'=N} k(\mathbf{\hat{x}_i,\hat{x}_{i'}})
 \end{split}
\end{equation}
Here, we use a linear combination of multiple RBF kernels ($k(\mathbf{x,\hat{x}})$) that is defined as :
\begin{equation}\label{eq:kernel}
 k(\mathbf{x,\hat{x}}) = \sum_{n}\eta_n \exp \left(\frac{-||\mathbf{x-\hat{x}}||^2}{2\sigma_n} \right)
\end{equation}

where $\sigma_n$ is the standard deviation and $\eta_n$ is the weight factor for $n^{th}$ RBF kernel.

The overall GAN loss for stage-2 is defined as:
\begin{equation}\label{eq:joint2}
    L_{GAN_2}(\mathbf{\theta_{G_2},\theta_{D_2},\theta_{R_2}})=L_{rec}+\alpha *L_{adv}+\beta * L_{reg} +\gamma* L_{Img}^{mmd}
\end{equation}
Here $\alpha$, $\beta$ and $\gamma$ are hyper-parameters. The architecture of this stage is similar to that of Stage-1, the only difference is that the generator $G_2$ takes input as $\mathbf{c,\hat{X_1}}$ i.e. the original attribute and the reconstructed sample.

\subsection{Stage-3}
This stage learns the joint embedding space between the generated image features from Stage-2 and the real image features based on class labels. This module consists of a Siamese Network which projects the real image and the synthesized image into a common subspace. The projection is made in such a way that the same class images are close to each other while the different class images are separated by a margin. Our ultimate goal is to generate image features based on the sketch such that the distribution of generated features should follow the same distribution as the real image features. However, this may not be true always since the domain shift may occur between the synthesized samples and the original samples. To reduce the domain shift, we project the data into a common space. In this module one network takes generated image features $\mathbf{\hat{X_2}}$ as input and second network takes real image features $\mathbf{X}$ as input and tries to learn a projection such that if the generated image feature $\mathbf{\hat{X_2}}$ and real image feature belongs to the same class, the similarity metric should be maximum. Otherwise, the similarity metric should be small. The loss function used in this module is as follows: 

First, we define true labels $\mathbf{Y_t}$
\[ \mathbf{Y_t} = \left\{ \begin{array}{ll}
         0 & \mbox{if $\mathbf{\hat{X_2}}$ and $\mathbf{X}$ belongs to the different class};\\
        1 & \mbox{if $\mathbf{\hat{X_2}}$ and $\mathbf{X}$ belongs to the same class}.\end{array} \right. \] 
False labels $\mathbf{Y_f}$ is defined as :
$\mathbf{Y_f}=1-\mathbf{Y_t}$

\begin{equation*}
    \mathbf{O_g}=NN_1(\mathbf{\hat{X_2}},\mathbf{\theta_N})
\end{equation*}
\begin{equation*}
    \mathbf{O_r}=NN_2(\mathbf{X,\theta_N})
\end{equation*}
\begin{equation*}
    \mathbf{d}=||\mathbf{O_r-O_g}||_2
\end{equation*}

\begin{equation}\label{eq:contrastive}
    L_C= \mathbf{Y_t} * d + \mathbf{Y_f} * (max[(m-d),0])^2
\end{equation}
Here $NN_{1}$ and $NN_{2}$ are the neural networks from the Siamese Network with shared weight. $\mathbf{\theta_{N}}$ is the parameter of the Siamese Network and $m$ is the margin hyper-parameter. The projected features $\mathbf{O_r}$ and $\mathbf{O_g}$ correspond to real image features and features generated from $G_2$ in stage-2 respectively. $\mathbf{O_r}$ and $\mathbf{O_g}$ are used for image retrieval task. $d$ is the Euclidean distance between $\mathbf{O_r}$ and $\mathbf{O_g}$.

\begin{table*}[t]
\begin{center}
\scalebox{0.83}{
\addtolength{\tabcolsep}{8pt}
\begin{tabular}{|l|l|c|c|c|c|}
\hline 
\textbf{Type} &  \multirow{2}{*}{\textbf{Method}} & \multicolumn{2}{c|}{\textbf{Sketchy Dataset}} & \multicolumn{2}{c|}{\textbf{TU Berlin Dataset}}\tabularnewline
\cline{3-6} 
& & \textbf{Precision@200} & \textbf{mAP@200} & \textbf{Precision@200} & \textbf{mAP@200}\tabularnewline
\hline 
\hline
& Baseline & 0.176 & 0.099 & 0.139 & 0.083\tabularnewline
& Siamese-1  \cite{siamese} & 0.243& 0.134 & 0.127 &0.061  \tabularnewline
& Siamese-2 \cite{sia2} & 0.251 & 0.149 & 0.133 &0.067 \tabularnewline
SBIR & Fine-Grained Triplet \cite{sketchy} & 0.155 & 0.081 & 0.086 & 0.050\tabularnewline
& Coarse-Grained Triplet \cite{CGT} & 0.169 & 0.083 & 0.128 &0.057 \tabularnewline
\hline
& Direct Regression & 0.066 & 0.022 & 0.117 &0.062 \tabularnewline
& ESZSL\cite{ESZSL} & 0.187 & 0.117 & 0.131 &0.072 \tabularnewline
& DAP \cite{lampert2014attribute}& 0.078 & 0.071 & 0.075 & 0.067\tabularnewline
ZSL-SBIR & SAE \cite{SAE2017}& 0.238 & 0.136 & 0.152 &0.084 \tabularnewline
& CAAE \cite{ashisheccv2018} & 0.240 & 0.146 & 0.159 & 0.094 \tabularnewline
& CVAE \cite{ashisheccv2018}&   0.269    &  0.159       &   0.182   &  0.109    \tabularnewline 
 & \textbf{SAN(Ours)}  & \textbf{0.322}& \textbf{0.236} & \textbf{0.218} & \textbf{0.141}\tabularnewline 
 \hline
 & CAAE \cite{ashisheccv2018}  &0.186 & 0.124 &0.162 & 0.0912 \tabularnewline 
GZSL-SBIR & CVAE\cite{ashisheccv2018} &0.202 & 0.134 & 0.177 &0.0985 \tabularnewline
& \textbf{SAN(Ours)} &\bf{0.304} & \textbf{0.227} & \textbf{0.203} & \textbf{0.124} \tabularnewline
\hline 
\end{tabular} 
}
\caption{Precision@200 and mAP@200 results on the traditional SBIR and ZSL method in the ZS-SBIR setup. Note that for a fair comparison, we reproduce the results using the same ResNet-152 features for all the baselines.  \cite{ashisheccv2018} proposed two models CAAE and CVAE.}
\label{tab:withoutside}
\end{center}
\end{table*}

\subsection*{Image retrieval methodology}
 During the test, we have sketches features of unseen classes. We aim to retrieve the same class images as sketches from an image database. Following are the steps involved in retrieving real images using sketches:
\begin{itemize}
    \item We pass the sketch features as the conditional variable $\mathbf{c}$ and a random vector $\mathbf{Z}$ to the trained generator $G_1$ which generates the corresponding image features $\mathbf{\hat{X_1}}$.
    \item The generated features $\mathbf{\hat{X_1}}$ along with its sketch features are passed to the trained generator $G_2$ which generates refined features $\mathbf{\hat{X_2}}$.
    \item Using trained Siamese network projected features $\mathbf{O_f}$ and $\mathbf{O_r}$ are obtained corresponding to the generated features $\mathbf{\hat{X_2}}$ and real image features $\mathbf{X}$ respectively.
    \item The real images are ranked according to the Euclidean distance $\mathbf{d(O_f, O_r)}$ for retrieval.
\end{itemize}
\section{Experiments Setting and Results}
\subsection{Dataset and Visual Feature}
We evaluate our proposed model on two widely used datasets for the task of ZS-SBIR:  Sketchy \cite{sketchy} and TU-Berlin \cite{berlin}, along with the additional images provided by the \cite{liu}.  Both the datasets are a collection of sketches and corresponding real images from several different categories. 

The visual features for images and sketches are extracted using ResNet-152 \cite{resnet} network pre-trained  on ImageNet-1000 dataset. No fine-tuning was performed. We forward pass the images and sketches in the pre-trained ResNet-152 model and extract 2048-dimensional features from the last fully connected layer. Visual features for the sketch is used as conditioning attributes for our proposed generative model.
\subsubsection{Sketchy Dataset(Extended)}\label{sketch}
The Sketchy dataset \cite{sketchy} contains sketch-image pairs from 125 different categories. Initially, there were 100 images from each category in the dataset. Hand-drawn sketches corresponding to the objects in these 12500 images were collected, resulting in 75471 sketches. Later \cite{liu} introduced 60502 more real images from all 125 classes resulting in a total of 73,002 images. We use a test-train split similar to \cite{ashisheccv2018} for the Sketchy dataset that contains 104 classes in the train set, and 21 classes in the test set. The split proposed by \cite{ashisheccv2018} ensures that none of the classes in the test set are present in the Imagenet-1000 classes. To form the sketch-image pair for training, we randomly select images and sketches from the same class and pair them. We make 1000 such pairs from each class to form the training set.

\subsubsection{TU Berlin Dataset(Extended)}
TU Berlin \cite{berlin} (extended) contains 250 different categories of sketches and images. It is a collection of 20000 sketches and 204489 images extended by \cite{liu,sketchnet}. We randomly select 30 classes for the test set and the remaining 220 classes for training. The dataset has some classes with large samples and some with only a few. To reduce the bias during training, we sample an equal number of sketches and images from each category. Following \cite{imagehashing}, during the test, we select only those classes with more than 400 samples. To form the image-sketch pairs for training, we follow the same strategy as the Sketchy dataset.

\subsection{Implementation details}
Our proposed network has following of 3 stages-\\
    \textbf{Stage-1}:
    Stage1 consists of a Generator, a Discriminator, and a Regressor Network. We use a series of fully connected (FC) layers in all these networks and apply ReLU after each layer except the last layer. A 300-dimensional noise vector $\mathbf{z}$, concatenated with a 2048-dimensional conditioning variable $\mathbf{c}$, is fed into the generator $G_1$. The conditioning variables $\mathbf{c}$ is a 2048-dimension features of sketches, obtained from ResNet-152  \cite{resnet}. $G_1$ passes the input features through a series of 4 FC layers having 1024, 512, 1024, 2048 neurons respectively, and outputs 2048-dimensional feature vector $\mathbf{\hat{X_1}}$ of the corresponding real image. Discriminator module $D_1$ tries to distinguish between the features of real images $\mathbf{X}$, and features generated $\mathbf{\hat{X_1}}$ from $G_1$. It takes 2048 dimension feature vectors and passes through a series of 3 FC layers having 1024, 512, and 128 neurons, respectively. It outputs the probability of the features being real. Regressor Network $R_1$ takes features generated from $G_1$ and tries to regenerate the features of the conditioning variable $\mathbf{c}$. It passes the input through a series of 4 FC layers having 1024, 512, 1024, and 2048 neurons, respectively. The output of the network is 2048-dimensional feature vector $\mathbf{\hat{c}}$.  We train our network using Adam Optimizer on  $L_{GAN_1}$ loss (Equation \ref{eq:joint1}) with learning rate = 0.00001, batch size = 50  keeping hyperparameters $\mathbf{\alpha}=$ 0.01 and $\mathbf{\beta}=$ 0.0001. We tune the $\mathbf{\alpha}$ and $\mathbf{\beta}$ hyper-parameters via a grid search from $10^-6$ to $10^3$. While training, we first train the discriminator separately for two epochs and then train the entire network end-to-end for $L_{GAN_1}$ loss. We observe that the validation performance saturates after 30 epochs.\\
 \textbf{Stage-2}:
    The network architecture of this stage is the same as that of Stage-1. The generator $G_2 $, of this stage, takes the output of $G_1$ concatenated with a conditioning variable $\mathbf{c}$ and outputs more refined features $\hat{X_2}$ closer to the real image features than the previous stage. This network is also trained using Adam Optimizer on  $L_{GAN_2}$ loss (Equation \ref{eq:joint2})  with learning rate = 0.00001, batch size = 50 keeping hyperparameters $\mathbf{\alpha}=$ 0.01, $\mathbf{\beta}=$ 0.0001 and $\mathbf{\gamma}=$ 0.01. We tune the $\mathbf{\alpha}$ $\mathbf{\beta}$ and $\mathbf{\gamma}$ hyper-parameters via a grid search from $10^-6$ to $10^3$. The training is done in a similar way, as described above for Stage1. We observe that the validation performance saturates after 35 epochs.\\
     \textbf{Stage-3}:
    This stage uses Siamese Network to find the similarity between the features generated in stage-2, namely, $\mathbf{\hat{X_2}}$ and the features of real image $\mathbf{X}$. It uses two similar neural networks $NN_1$ and $NN_2$ with shared weights to process both the input features.  $NN_1$ and $NN_2$ has an input FC layer with 1024 neurons followed by a ReLU layer and an output FC layer with two neurons. We minimize the contrastive divergence loss between $\mathbf{X_g}$ and $\mathbf{X_r}$ features obtained by passing input features $\mathbf{\hat{X_2}}$ and $\mathbf{X}$ through $NN_1$ and $NN_2$ respectively. We train the network using Adam optimizer on the contrastive divergence loss $L_C$ (Equation \ref{eq:contrastive}) setting hyperparameter $\mathbf{m}$ = 5 with learning rate 0.01 and batch size 32. We tune the hyper-parameter $\mathbf{m}$ via a grid search from $1$ to $100$. We train the network for 20 epochs and observe that the validation performance saturates after 15 epochs.

\begin{figure}[t!]
\centering
\includegraphics[scale=0.175]{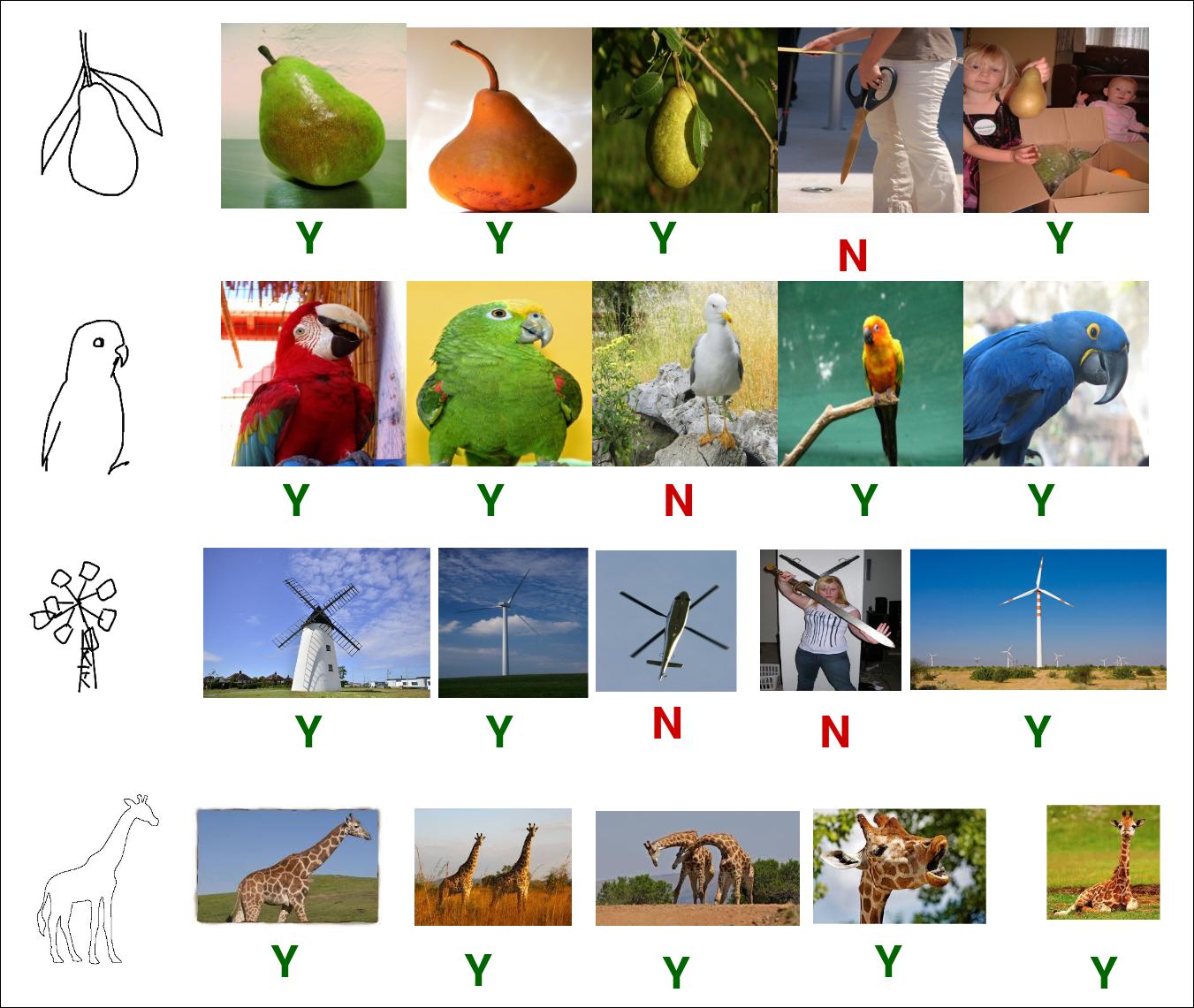}
\caption{Top 5 Retrieval results of our proposed model. Here we can see that a  retrieved object fails when the sketch outline is very close to the image outline. $\mathbf{N}$ indicates false-positive retrieval results.}
\label{fig:result}
\end{figure}

\subsection{Comparison with existing methods}
We compare our proposed model with the existing state-of-the-art of SBIR, ZSL baselines, and recently proposed ZS-SBIR approaches.
\subsubsection{Comparison with SBIR baseline}
The baseline models of SBIR includes Siamese-1\cite{siamese}, Siamese-2\cite{sia2}, Fine-Grain Triplet(FGT)\cite{sketchy} and Coarse-grained triplet(CGT)\cite{CGT}.
All the models were built according to the description in the original paper and trained under the zero-shot setting. We use the same seen-unseen splits of categories for all the experiments for a fair comparison. A baseline also added for comparison. We take a ResNet-152 network pre-trained on ImageNet-1K as the baseline. The score for a given sketch-image pair is given by the cosine similarity between their ResNet-152 features.
\subsubsection{Comparison with ZSL baseline}
We select a set of state-of-art zero-shot learning approaches as the benchmark and implement the same for the sketch-based image retrieval task. The selected ZSL algorithms involves Direct Regression, ESZSL\cite{ESZSL},  DAP\cite{lampert2014attribute}, and SAE\cite{SAE2017}. The Semantic Autoencoder (SAE) proposes an autoencoder framework to encourage the re-constructibility of the sketch vector from the generated image vector. ESZSL\cite{ESZSL} learns a bilinear compatibility matrix between images and attribute vectors in the context of zero-shot classification.
We adapt the model to the ZS-SBIR task by mapping the sketch features to the image features using labeled training data from the seen classes. In Direct-Regression, the ZS-SBIR task is formulated as a simple regression problem where each image feature vector is predicted from the sketch features. This is similar to the direct attribute prediction method that is a widely used baseline for zero-shot image classification. 

\subsubsection{Comparison with ZS-SBIR }
Recently ZSIH \cite{imagehashing}, CVAE \cite{ashisheccv2018} and Doodle to search \cite{doodle} methods are proposed for ZS-SBIR.  Both these methods \cite{doodle,imagehashing} use side information(word vector \cite{glove} for sketch classes) along with sketch features to train the model. \cite{ashisheccv2018} proposed two generative models, CVAE \cite{ashisheccv2018} and CAAE \cite{ashisheccv2018} that use only sketches features as a condition to synthesize image features(without using any side information). CAVE\cite{ashisheccv2018} and CAAE\cite{ashisheccv2018} have performed experiments only on the sketchy dataset in a new split of seen and unseen classes, whereas ZSIH \cite{imagehashing} and Doodle to search \cite{doodle} have shown experiments on both the Berlin and Sketchy datasets. However, all these methods have shown experiments only in the standard zero-shot setting(ZSL). So, for a fair comparison, we compare our proposed model with CVAE\cite{ashisheccv2018} and CAAE\cite{ashisheccv2018}. 

\subsection{Results and Analysis}
From Table \ref{tab:withoutside}, we observe that all the SBIR and ZSL baselines are not able to generalize well for unseen class sketches. The reason for their failure is that these methods have been trained in a supervised setting and hence have not used any transfer learning techniques for unseen classes.

For a fair comparison we reproduce the results of CVAE\cite{ashisheccv2018} and CAAE\cite{ashisheccv2018} for ResNet-152 features on Sketchy(on realistic split) and Berlin(on random split) dataset. We perform experiments in both standard and generalized ZSL settings. We observe that in Standard ZSL setting, our model outperforms CVAE by $5.3\%$, $7.7\%$, and $3.6\%$, $3.2\%$ absolute improvement in precision@200 and mAP@200 in Sketchy and Berlin dataset respectively. For GZSL, we randomly sampled $10\%$ examples per class from seen classes and included with unseen class examples to create test data for our proposed model. Our model outperforms CVAE by $10.2\%$, $9.3\%$, and $2.6\%$, $2.5\%$ absolute improvement in precision@200 and mAP@200 in Sketchy and Berlin dataset respectively. We observe that our model without using any side information outperforms the Doodle to search\cite{doodle} in the Berlin dataset that uses the sketch class description as side information to train the model.

Figure \ref{fig:result} shows the top-5 retrieval results of our model for sketches of unseen classes. The retrieved images show that our proposed approach is robust for unseen classes, and it learns a better mapping from sketch space to image space.
\section{Ablation Analysis}
In this section, we show some ablation studies to prove the plausibility of our proposed model. Tables \ref{tab:sketchy} and \ref{tab:newsplit1} clearly show the significance of each stage in our proposed model.

\begin{table}[htb!]
\begin{center}
\scalebox{.75}{
\begin{tabular}{|l|l|c|c|}
\hline
\multicolumn{4}{|c|}{\textbf{Berlin Dataset}} \tabularnewline
 \hline 
 \textbf{Type}&{\textbf{Method}} & \multicolumn{1}{c|}{\textbf{Precision@200}} & \multicolumn{1}{c|}{\textbf{mAP@200}}\tabularnewline
\cline{2-3} 
\hline 
\hline
{$G_1$ } & SAN-SBIR(our) &0.170 &  0.101\tabularnewline 
{$G_1+G_2$ } & SAN-SBIR(our) &0.191 &  0.119\tabularnewline \hline
{\bf{Improvement}} & \bf{With $G_2$}  &\bf{2.1\%} &  \bf{1.8\%}\tabularnewline 
\hline
{$G_1+G_2$ } & SAN-SBIR(our) &0.191 &  0.119\tabularnewline
{$G_1+G_2$ + MMD} & SAN-SBIR(our) &0.204 &  0.128\tabularnewline \hline
{\bf{Improvement}} & \bf{With MMD}  &\bf{1.3\%} &  \bf{0.9\%}\tabularnewline 
\hline
{$G_1+G_2$ + MMD} & SAN-SBIR(our) &0.204 &  0.128\tabularnewline
{$G_1+G_2+MMD+P$} & SAN-SBIR(our) &0.218 &  0.141\tabularnewline
\hline
{\bf{Improvement}} & \bf{With P}  &\bf{1.4\%} &  \bf{1.3\%}\tabularnewline 
\hline
\end{tabular}  
}
\caption{Precision@200 and mAP@200 results of our proposed approach on ZS-SBIR setup for Berlin Dataset. $\mathbf{G_1,G_2,P,MMD}$ corresponds to Stage-1, Stage-2, Stage-3 and maximum mean discrepancy respectively.}
\label{tab:sketchy}
\end{center}
\end{table}

\begin{table}[htb!]
\begin{center}
\scalebox{.75}{
\begin{tabular}{|l|l|c|c|}
\hline
\multicolumn{4}{|c|}{\textbf{Sketchy Dataset}} \tabularnewline
 \hline 
 \textbf{Type}&{\textbf{Method}} & \multicolumn{1}{c|}{\textbf{Precision@200}} & \multicolumn{1}{c|}{\textbf{mAP@200}}\tabularnewline
\cline{2-3} 
\hline 
\hline
{$G_1$} & SAN-SBIR(our) &0.284 &  0.189\tabularnewline 
{$G_1+G_2$ } & SAN-SBIR(our) &0.297 &  0.205\tabularnewline \hline
{\bf{Improvement}} &  \bf{With $G_2$} &\bf{1.3\%} &  \bf{1.6 \%}\tabularnewline 
\hline
{$G_1+G_2$ } & SAN-SBIR(our) &0.297 &  0.205\tabularnewline
{$G_1+G_2$ +MMd} & SAN-SBIR(our) &0.314 & 0.218\tabularnewline
\hline
{\bf{Improvement}} &  \bf{With MMD} &\bf{1.7\%} &  \bf{1.3 \%}\tabularnewline 
\hline
{$G_1+G_2$ +MMD} & SAN-SBIR(our) &0.314 & 0.218\tabularnewline 
{$G_1+G_2+MMD+P$} & SAN-SBIR(our) &0.322 & 0.236\tabularnewline
\hline 
{\bf{Improvement}} & \bf{With P} &\bf{0.8\%} &  \bf{1.8\%}\tabularnewline
\hline 
\end{tabular}  
}
\caption{Precision@200 and mAP@200 results of our proposed approach on ZS-SBIR setup for Sketchy dataset. $\mathbf{G_1,G_2,P,MMD}$ correspond to Stage-1, Stage-2, Stage-3 and maximum mean discrepancy respectively.}
\label{tab:newsplit1}
\end{center}
\end{table}


\subsection*{Ablation with multi-stage GAN}
 Our model generates more robust features with two stages $G_1+G_2$  for unseen classes based on sketches, the improvement of performance in $G_1+G_2$ over $G_1$ justifies our claim. With $G_1+G_2$ there is $2.1\%$, $1.8\%$ and $1.3\%$, $1.6\%$ absolute performance improvement in precision@200 and mAP@200 for Berlin and Sketchy datasets respectively as compared to only $G_1$.
 
 \subsection*{Effect of MMD Loss}
Our ablation shows that adding MMD loss in the Generator of the second stage ($G_2)$ has boosted the model performance. The MMD loss enforces the model to maximize the margin between generated samples of a different class, therefore increases the robustness of the retrieval task. We found an absolute improvement of  $1.3\%$, $0.9\%$ and $1.7\%$, $1.3\%$  in precision@200, and mAP@200 as compare to without using MMD for Berlin and Sketchy datasets respectively.
 
\subsection*{Ablation with Siamese Network}
\begin{figure}[t]
    \centering
    \includegraphics[scale=0.23]{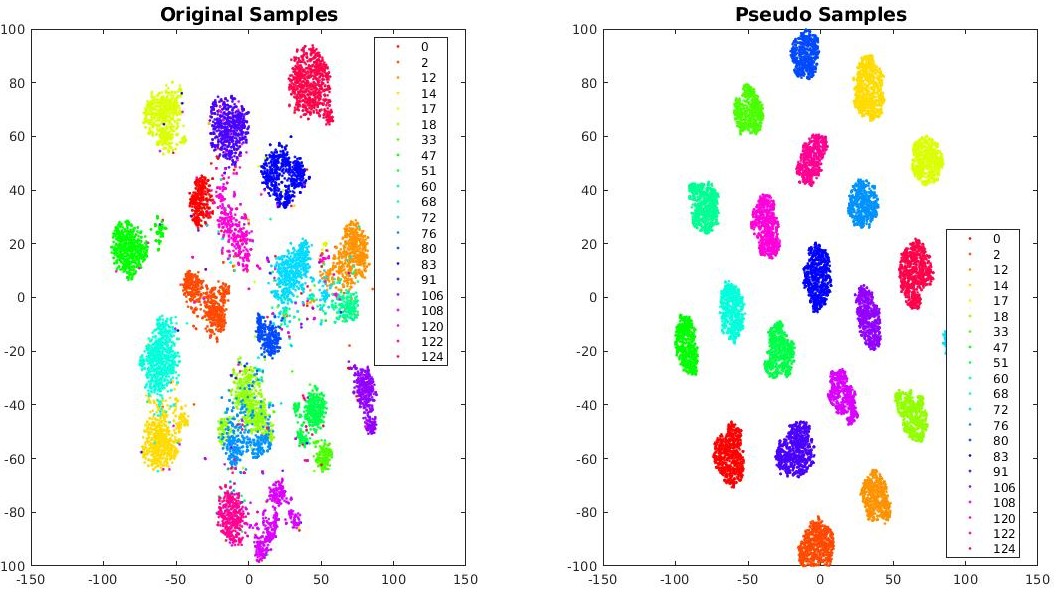}
    \caption{tSNE-Visualization of the original and synthesized samples. We can see the generated samples follow the same distribution as the original one, and the projected features for unseen classes are discriminative and class-wise well separated.}
    \label{fig:tsne}
\end{figure}

Hubness may occur on applying the nearest neighbor search on generated features for the task of image retrieval that may degrade the performance of our model. \cite{hubness} shows that the probability of becoming a hub node is high if we compute the KNN in the original space, whereas if we compute it in a mapped space, the hubness problem reduces as compared to previous one. We address this issue in stage-3 (transformation stage) of our model. In this stage, the features generated by stage2 and the real image features are projected to a common space by similarity using a Siamese Network. The projection is made such that the features of the same class are close, while a significant margin separates the features of different classes. This approach provides more class-wise discriminative features. Tables \ref{tab:sketchy} and \ref{tab:newsplit1}   do establish that the inclusion of stage-3 does improve performance significantly. Including stage-3, the absolute performance of our model improves by $1.4\%$, $1.3\%$ and $0.8\%$, $1.8\%$  in precision@200 and mAP@200 as compare to 2 stage model ($G_1+G_2+MMD$) for Berlin and Sketchy datasets respectively. Figure \ref{fig:tsne} shows the tSNE visualization of original features and synthesized features in projected space, and we can observe that the projected features are well class-wise separated. 
 
\section{Conclusion}
In this paper, we propose to use a multi-stage GAN based framework called SAN to solve the SBIR problem in a zero-shot setting.  The proposed approach uses SAN to synthesize refined image samples from the sketch features and hence reduces the SBIR problem to an image-to-image retrieval problem. The proposed method is based on a multi-stage GAN to synthesize refined samples. The nearest neighbor search technique for the SBIR task suffers from the hubness \cite{hubness} problem. To address this issue, we project the data to another space using Siamese Network, where hubness has its minimal effect. In the ablation study, we found that all the proposed components (Stage-1, Stage-2, Stage-3) have a significant contribution to improving the performance of the ZS-SBIR task. We perform an extensive experiment on Sketchy and TU-Berlin datasets for the ZS-SBIR in both ZSL and GZSL settings. Our proposed approach shows the state-of-the-art result without using any additional information to train the model.

{\small
\bibliographystyle{ieee}
\bibliography{egbib}
}

\end{document}